\def\year{2021}\relax
\documentclass[letterpaper]{article} % DO NOT CHANGE THIS
\usepackage{aaai21}  % DO NOT CHANGE THIS
\usepackage{times}  % DO NOT CHANGE THIS
\usepackage{helvet} % DO NOT CHANGE THIS
\usepackage{courier}  % DO NOT CHANGE THIS
\usepackage[hyphens]{url}  % DO NOT CHANGE THIS
\usepackage{graphicx} % DO NOT CHANGE THIS
\usepackage{natbib}  % DO NOT CHANGE THIS AND DO NOT ADD ANY OPTIONS TO IT
\usepackage{caption} % DO NOT CHANGE THIS AND DO NOT ADD ANY OPTIONS TO IT
\usepackage{amsmath}
\usepackage{mathtools}
\usepackage{amssymb}
\usepackage[symbol]{footmisc}

\usepackage{caption}
\usepackage{subcaption}

\title{Multimodal Sensory Learning for Real-time, Adaptive Manipulation}
\author{
    Ahalya Prabhakar\thanks{Corresponding Author}, Stanislas Furrer\thanks{Authors contributed equally}, Lorenzo Panchetti\footnotemark[2], Maxence Perret\footnotemark[2], Aude Billard\\
}
\affiliations{
    Ecole Polytechnique Federal de Lausanne\\
    CH-1015 Lausanne, Switzerland\\
    \{ahalya.prabhakar, stanislas.furrer, lorenzo.panchetti, maxence.perret, aude.billard\}@epfl.ch
}

\begin{document}

\maketitle

\begin{abstract}
Adaptive control for real-time manipulation requires quick estimation and prediction of object properties. While robot learning in this area primarily focuses on using vision, many tasks cannot rely on vision due to object occlusion. Here, we formulate a learning framework that uses multimodal sensory fusion of tactile and audio data in order to quickly characterize and predict an object's properties. The predictions are used in a developed reactive controller to adapt the grip on the object to compensate for the predicted inertial forces experienced during motion. Drawing inspiration from how humans interact with objects, we propose an experimental setup from which we can understand how to best utilize different sensory signals and actively interact with and manipulate objects to quickly learn their object properties for safe manipulation. 
\end{abstract}

\section{Introduction}

\begin{figure}
    \centering
    \includegraphics[width=0.5\textwidth]{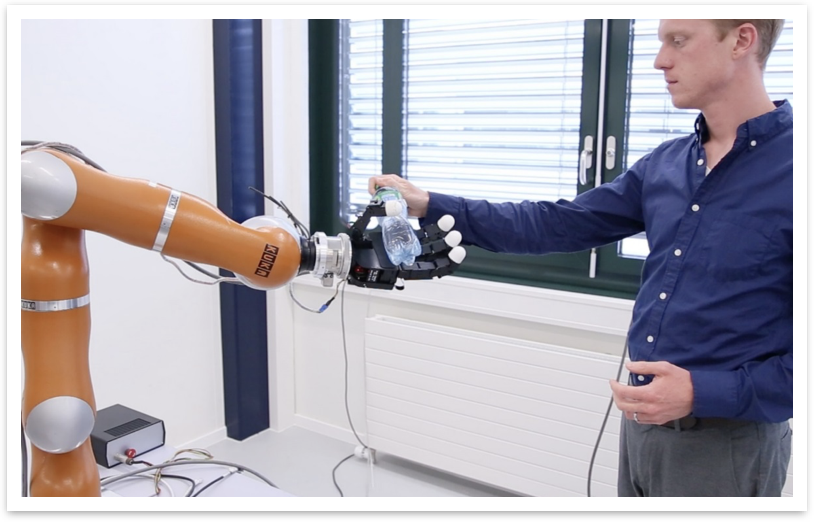}
    \caption{Robot Handover. Real-world object handover requires the ability for the robot to quickly estimate the physical properties of the object for safe manipulation \cite{7803296}. }
    \label{fig:robot_handover}
  \end{figure}

Efficient learning about the physical properties of the world and the objects in it is necessary during human-robot interaction tasks, where the robot may need to quickly learn and communicate information about the objects to the user. For example, in a setting where a robot is assisting a human in reaching for and handing over a container with specific contents (shown in Fig. \ref{fig:robot_handover}, the robot must be able to interact with objects and quickly estimate their contents and physical properties in it. During collaborative tasks in a workplace or home, the robot may need to assist the person by quickly handing over specific objects in a safe manner. Doing so requires both the ability to quickly estimate the object's contents, and, more critically, physical properties to ensure the handover is done in a safe manner. Being able to model the inertial properties of the objects enables the robot to anticipate high torques and forces exerted by the object, which can reduce the impact on the robot's joints. Furthermore, even if the robot can handle higher forces, it can minimize the inertial forces exerted during handover, such that the user receiving the object can do so in a safe manner, without dropping the object or injuring themselves in the process. 

% Quickly developing an estimation of the physical properties of the world and the objects in it is a critical skill for successful manipulation in daily tasks. 
Humans are skilled at integrating experienced sensory signals in order to efficiently construct a perceptual model of the physical world. This can require fusing and processing high-dimensional sensory information quickly into a combined low-dimensional representations in order to build our understanding of the physical world. When one sensory modality is incapable of providing the necessary information (e.g., vision due to occlusion), we compensate with other sensory information, typically tactile or proprioceptive data to estimate physical properties \cite{ernst2002humans}.  To do so, we interact with the world in intentional ways in order to most efficiently maximize information from multiple sensory signals (e.g., actively manipulate object to maximize the information from audio and tactile signals) \cite{ERNST2004162}.

Sensory perception in robotics, however, still relies on large amounts of data to learn and can take a long time to process and learn from \cite{eslami2018neural, NIPS2015_d09bf415}. As such, processing the sensor data to enable real-time adaptation is still a challenge, which is particularly critical for robotic manipulation tasks. Furthermore, most model learning primarily focuses on vision sensing \cite{metta2003early, schiebener2013integrating, katz2008manipulating}. However, for many tasks in our daily lives (e.g., retrieving an object from the fridge), the physical properties of the object may not be estimated from vision alone due to occlusion and opacity of the container (e.g., a carton of milk). 

\begin{figure*}[bth]
    \centering
    \includegraphics[width=0.9\textwidth]{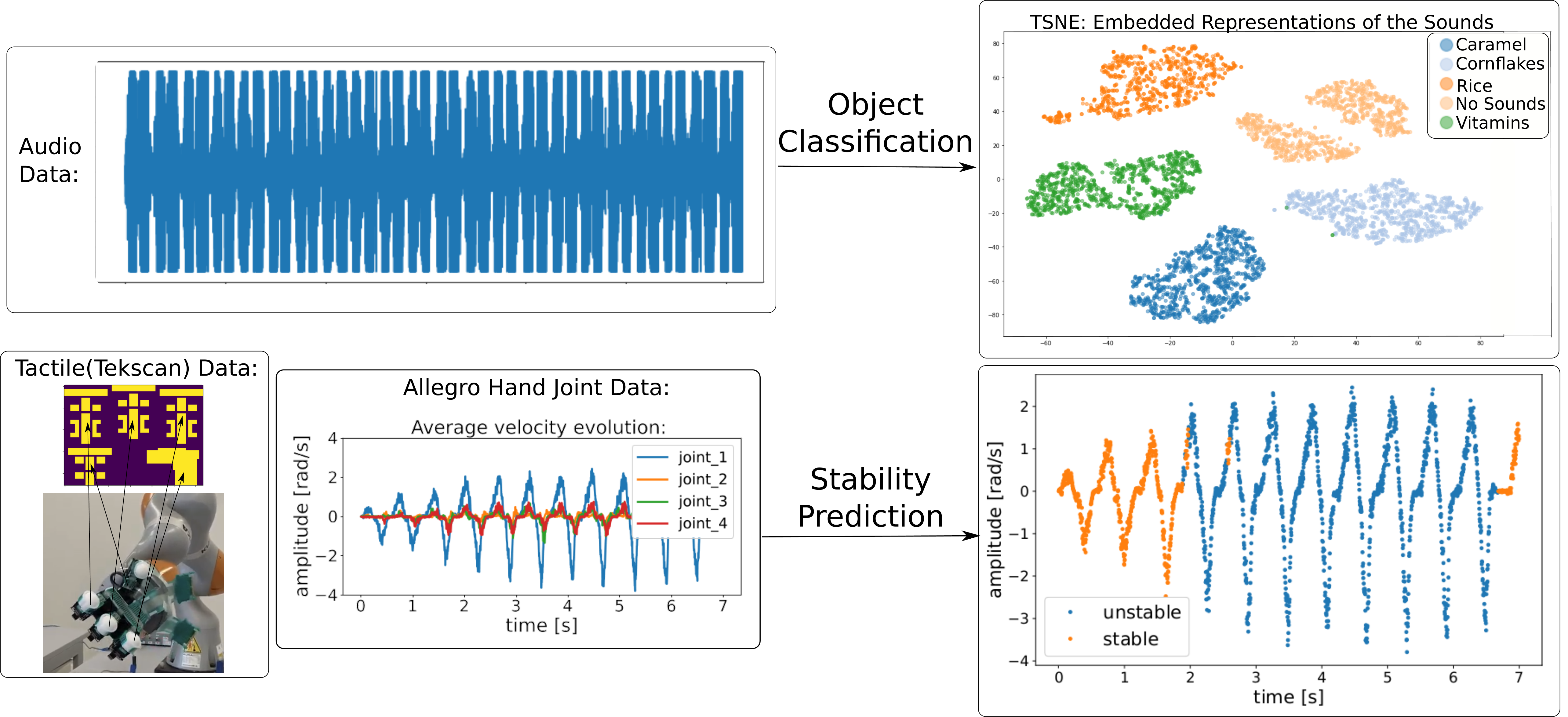}
    \caption[Data Learning]{Data used for Learning and Realtime Prediction.  The audio data is used to predict the material contents inside the container. The tactile and force-torque data is used for predicting grip stability. }
    \label{fig:data_learning}
  \end{figure*}
Here, we focus on this scenario of learning and quickly estimating the inertial properties of the manipulated objects when vision is not available. We instead rely on fusing multimodal sensory information from tactile and audio sensing, two sensory modalities often underutilized in robotics. The challenge of learning and predicting the physical properties is exacerbated by the need to predict the inertial properties quickly enough to enable real-time adaptive control for maintaining stable grasp of the object--- on the order of milliseconds to be feasible. As such, another key challenge is intentionally controlling and manipulating the object based on the information gain to enable efficiently estimating the object properties. By actively manipulating the object, we quickly predict properties (e.g., mass and inertia during shaking) to ensure manipulation without dropping the object.

The main contributions of this proposed work are as follows: 1) We develop a framework for learning and predicting the contents and inertial properties of objects inside a container using multi-modal sensory fusion. 2) Using the predictions, we enable real-time adaptive control of the object to minimize the effort and torque experienced on the joints of the robot hand. 3) We develop an active learning hierarchical framework that intentionally manipulates the objects with specific motions in order to enable efficient learning of the object properties. In addition to the main scientific contributions of the work, we are generating a manipulation dataset with audio and tactile sensory data for manipulating different objects that is available to the public \footnote[4]{Data is available at the link: http://www.epfl.ch/labs/lasa/datasets/}.  

\section{Related Work}
Learning object properties and contact force estimation for in-hand manipulation has been a long-studied research topic. Some work has focused on learning and estimating object properties to enable accurate manipulation \cite{wettels2011haptic, DeMariaFrictionEstimation, DensePhysNet, yuan2017connecting}.  Similar to our approach, some research in force estimation has focused on slip detection \cite{chen2018tactile, veiga2018grip, agriomallos2018slippage, heyneman2016slip} and have extended it to incorporate force prediction \cite{SuForceEstimation2015, Ma2019} for enabling feedback-control methods. 

To improve force estimation, some work has utilized multimodal sensory fusion to classify objects and better estimate object properties. However, most research doing so have focused on combining vision and tactile sensory data to do so \cite{kragic2005vision, Hasan2020, yamaguchi2016combining, Zhang2019, sundaralingam2019robust}. More recently, the benefits of auditory sensing has been explored for robot manipulation for object classification \cite{sinapov2011interactive, Mir2021, guler2014s} and property estimation \cite{Eppe2018, Gandhi-RSS-20, jonetzko2020}. In this work, we utilize similar approaches of combining auditory and tactile data in order to improve contact force prediction for improved grip control in real-time settings. 
\section{Methods}
  
\subsection{Hardware Pipeline}

\begin{figure}[bth]
    \centering
    \includegraphics[width=0.3\textwidth]{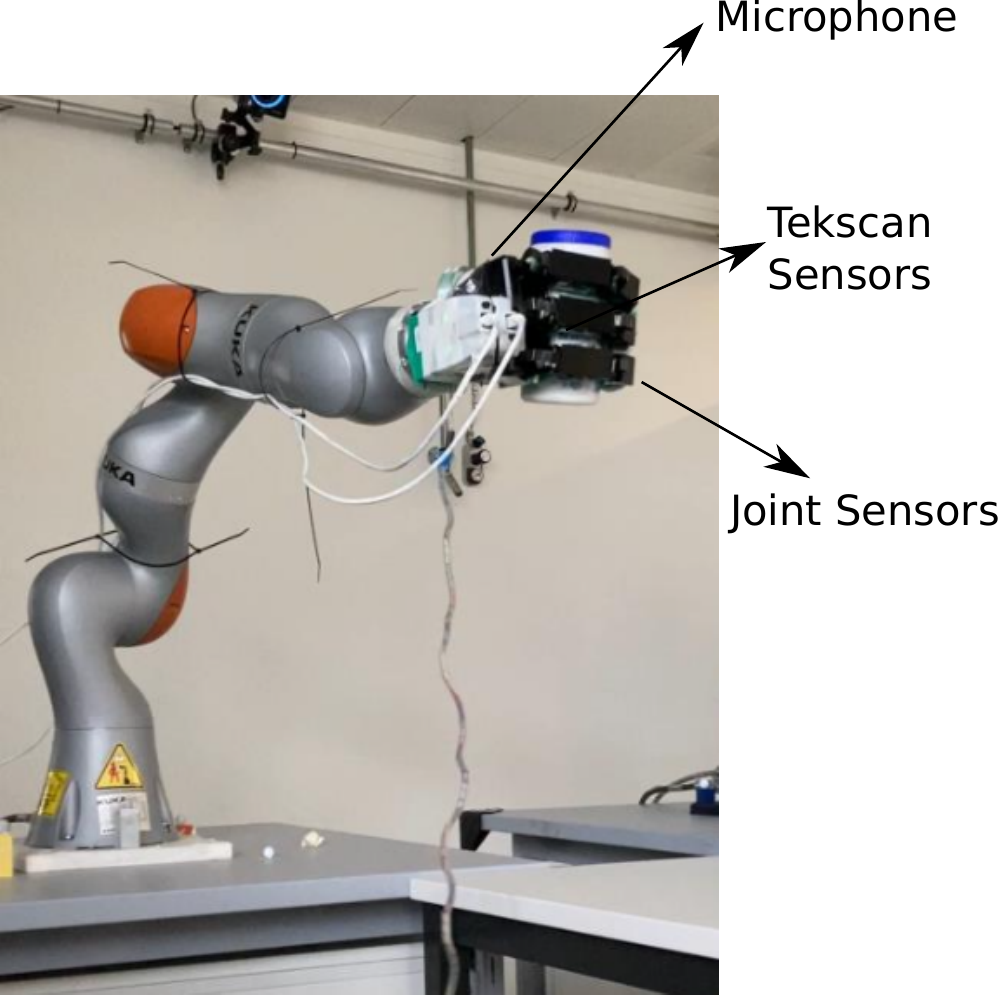}
    \caption[Experimental Setup]{The full experimental setup, which includes an Allegro hand mounted on a Kuka IIWA 7 robot arm. Tekscan tactile sensors cover the allegro hand surface provides pressure data. A microphone is attached to the hand to record the audio of the objects moving inside the container. }
    \label{fig:experimental_setup}
  \end{figure}
The experimental setup, shown in Figure \ref{fig:experimental_setup}, uses a Kuka IIWA 7 with an attached Allegro hand for robotic manipulation. An audio microphone is attached on the hand, to capture the sounds of the contents inside the container being manipulated. A Tekscan tactile sensor covering the Allegro hand provides tactile pressure data across the hand. Finally, the joint data of the Allegro hand is also captured. 
\begin{figure*}[hbt]
    \centering
    \includegraphics[width=0.7\textwidth]{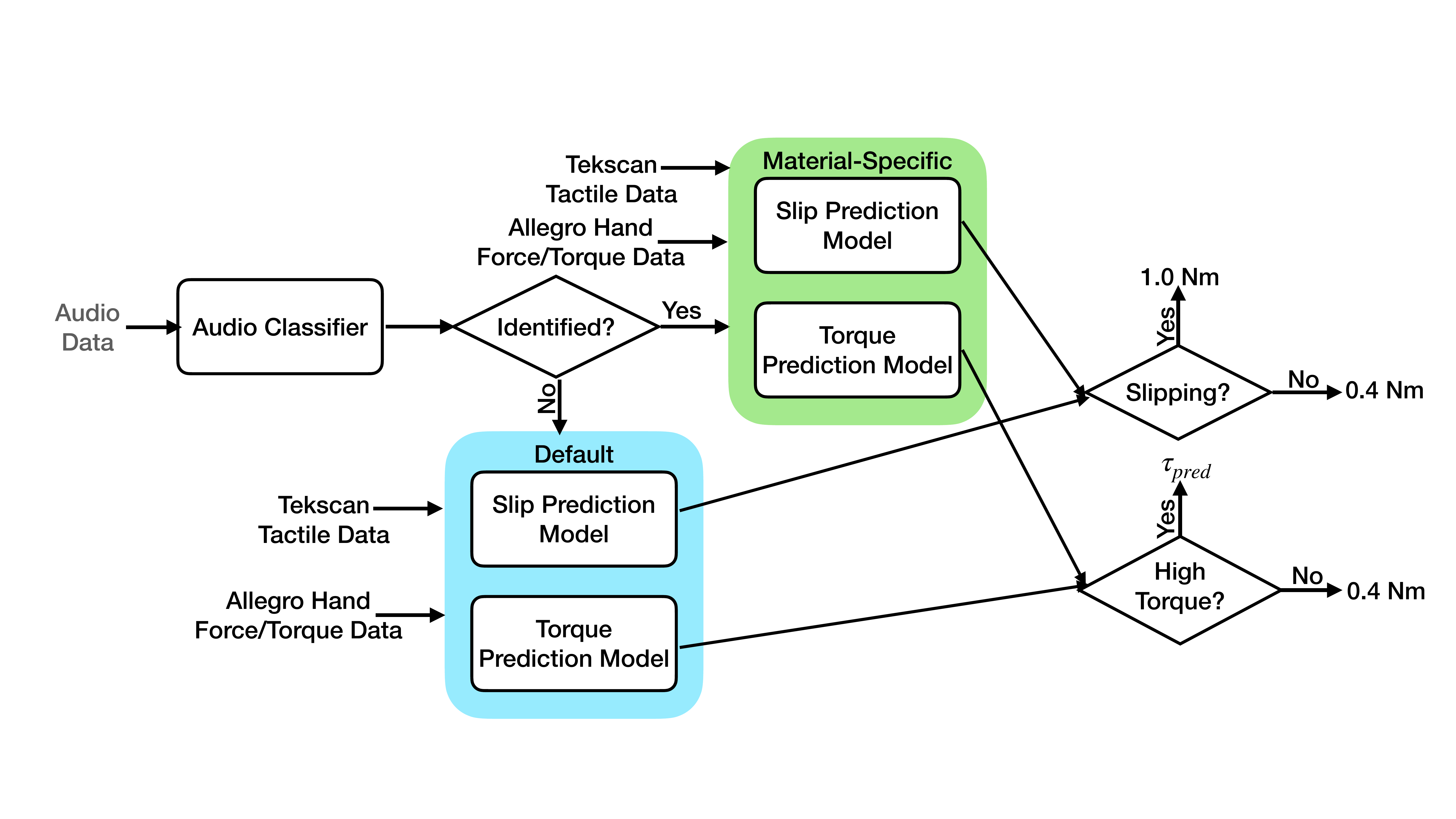}
    \caption[Full Pipeline]{The full robotic pipeline with all the different learning and prediction components integrated. The audio data is used to predict the material contents inside the container. The tactile and force-torque data is used for slip and torque prediction to enable real-time reactive control of the Allegro hand to maintain stable grip on the container with minimal effort. }
    \label{fig:nn_pipeline}
  \end{figure*}
\subsection{Experimental Setup}
We test 4 different types of objects--- rice, cereal, gummy candies, and vitamins. We use the same volume (1 cup) of each of the materials inside the container, which is an opaque plastic bottle. 

\subsection{Robot Control}
We create a library of two motion types to manipulate the object--- vertical shaking and rotation. In order to ensure sufficient speed and acceleration to generate noise during the manipulation process, we first used the Optitrack motion capture system to record a human manipulating the container in the different ways. Using the captured data, we generate velocity and acceleration profiles for each of the motion types. For the shaking motions, the acceleration-based control system uses a passive DS controller \cite{kronander2015passive} to maintain position in the center location, and then additionally applies torque directly to specific joints such that the acceleration matches the captured acceleration profiles to generate audio during motion. This shaking torque is applied to shake the container a desired number of times. The rotation motion uses a position-based DS controller that combines a DS control to maintain the desired position, while changing the orientation a specified range in a sinusoidal pattern to generate the motion. 

\subsection{Prediction Learning}

 \subsubsection{Data Collection} 
The learning framework consists of two separate neural networks. The audio data is used to predict the object's contents. The audio data is trained on 5 classes- the four materials (rice, cereal, gummies, and vitamins) and a emmty class with no contents in the container. 
For data acquisition, we generate 30 trials of each motion with each material, a total of 10 different classes. 

\subsubsection{Network Setup and Training}
\label{sec:network_setup}
The audio data is used to enable fast material classification during manipulation.  The tactile data from the Tekscan sensors and the force-torque data from the Allegro hand joints are used to predict slip on the container and predict torque that will be experienced, given knowledge of which motion type is being executed. For material classification, the audio data is first preprocessed to crop the audio clip to the start and end of the motions. To improve the speed of classification, the audio data is the segmented into a set of shorter 1-second segments. The segmented data is then augmented using a combination of pitch shifting and adding noise. The augmented dataset is then processed to extract the Mel-frequency cepstral coefficients (MFCCs) of each audio segment. The MFCC data is input into a CNN classification model to classify the material in the audio segment. 

For slip prediction, we use haptic features from the tekscan sensor data--- specifically the mean and maximum values of non-zero forces, center of mass and gradient of the center of mass experienced across the palm--- as well as the value and the change in allegro hand joint angles. Using these in an RNN regression and classifier model, we forward predict the value and location of maximum force that will be experienced, as well as if slip is occurring. We classify slip as occurring by a thresholded change in Allegro hand joint positions. This threshold, currently manually specified based on qualitative observations, will be quantitatively specified using a vision tracking system on the container to observe its position relative to the hand and how it changes during slipping. 

Figure \ref{fig:data_learning} shows preliminary example results of the outputs of the different classification networks using datasets collected manually. For the audio data, the example audio data and the t-SNE plot result from an audio classifier trained and tested on a preliminary dataset collected by manually shaking the container. The tactile results shown use a slip detection classifier trained by manually shaking the allegro hand holding the container, and testing with the robot performing the motions by using the Optitrack motion tracker to follow a user performing the same motions. These preliminary results are promising, though using the dataset collected with the robot systematically executing the different motion types with all the integrated sensors mounted on the robot should improve real-time performance on the robot.

\subsection{Reactive Controller}

The audio predictions and both the torque and slip tactile predictions will be used in the real-time reactive controller for the robotic hand. From the data, we will train slip and torque prediction models. Default prediction models will be learned over all the data collected to predict slip and torque experienced, for each motion type. Additionally, material-specific models will be learned from each material data for each motion type. During real-time execution, the default model will be initialized according to the specific motion. Once the audio classifier predicts the contents of the container, the corresponding material-specific slip prediction models will be used. 

The Allegro hand controller uses a power grasp hold on the object during the motion at standard 0.4 Nm of torque. Though the maximum grasp torque applied can be 1.0Nm, higher torque application quickly leads to motor burnout, highlighting the importance of adaptive control during manipulation. The controller uses the slip and torque prediction models described in Section \ref{sec:network_setup} to reactively adapt the grasp control. If the slip model predicts slipping, the torque applied increases until the grasp is stable and slip is not predicted. Furthermore, the torque prediction is used to stiffen the torques on the joint when high forces are predicted to be experienced due to the objects' inertial properties as they move inside the container. The full integrated pipeline is shown in Figure \ref{fig:nn_pipeline}.

\section{Planned Experiments and Future Work}
The training and testing of the full pipeline with all the learning components will be performed to demonstrate successful learning and real-time prediction of an object's inertial properties during manipulation. The results will be analyzed to evaluate learning success as well as to learn about the information content of the different types of sensory signals. In particular, the force-torque data and the pressure data will be compared to understand the use of the different types of tactile data for informing real-time slip and torque prediction during manipulation.  Furthermore, the different motion types will be analyzed for determining information gain for content prediction in order to generate a hierarchical Bayesian inference control framework that exploits the information content of the different motion types to enable efficient learning and prediction of the object's contents and properties. This will be used to develop a adaptive controller that anticipates the inertial forces that will be experienced and and adapts the torque controls to minimize slip during handover to a user. 

\section{Conclusion}
In this work, we propose an approach that uses multimodal sensory fusion to enable real-time learning and prediction of an object's inertial properties for adaptive manipulation. Using inspiration from human manipulation, we propose utilizing tactile and auditory feedback to enable efficient learning through intentional manipulation. By evaluating the learning from the sensory signals and motions, we plan to evaluate the information content and quality of the different sensory signals and the different motions. Using this analysis, we plan to develop an active learning framework that relies on Bayesian inference to intentionally choose the specific motions to enable efficient learning, a requirement when needing to interact in the real-world in human-robot collaboration settings. 

\section{ Acknowledgments}
This work was supported by the European Research Council by the CHIST-ERA program through the project CORSMAL and through the SAHR Project under the ERC Advanced Grant 741945.
\bibliography{multimodalmanipulation-references.bib}
\end{document}